%% file: main.tex
\documentclass[sigconf,nonacm]{acmart}

\usepackage{booktabs}
\usepackage{subcaption}
\usepackage{algorithm}
\usepackage{algorithmic}
\usepackage{float}   
\usepackage{enumitem}
\usepackage{tikz}
\usetikzlibrary{calc,positioning,fit,arrows.meta}

\definecolor{skInk}{HTML}{1F2937}
\definecolor{skAlign}{HTML}{2F5D8C}
\definecolor{skAlignBg}{HTML}{E8EFF7}
\definecolor{skQual}{HTML}{2E7D6B}
\definecolor{skQualBg}{HTML}{E6F2EE}
\definecolor{skWarn}{HTML}{C1543A}
\definecolor{skWarnBg}{HTML}{FBEAE5}
\definecolor{skGray}{HTML}{8A939F}
\definecolor{skGrayBg}{HTML}{F2F4F6}

\newcommand{\skillbox}[3]{%
  \begin{tikzpicture}
    \node[anchor=north west, inner sep=0pt] (sk@body) at (0,0)
      {\begin{minipage}{\dimexpr\columnwidth-14pt\relax}
         \ttfamily\fontsize{5.5}{6.6}\selectfont\color{skInk}#3
       \end{minipage}};
    \node[fit=(sk@body), inner xsep=5pt, inner ysep=4.5pt,
          rounded corners=2pt, draw=#1!70, line width=0.5pt] (sk@frame) {};
    \node[anchor=south west, rounded corners=2pt, fill=#1, text=white,
          inner xsep=5pt, inner ysep=2.6pt,
          minimum width=\dimexpr\columnwidth-4pt\relax, align=left]
      at ([yshift=1.6pt]sk@frame.north west)
      {\sffamily\fontsize{6.4}{6.4}\selectfont\bfseries #2};
  \end{tikzpicture}%
}
\setlist[itemize]{leftmargin=*,labelsep=0.4em,topsep=3pt,itemsep=2pt,parsep=0pt}

\newcommand{\datasetname}{PROSE}

\begin{document}

\title{SAGE: Self-Evolving Storyboard Skills via Attribution-Guided Rule Evolution}

\author{Maolin Ran}
\affiliation{%
  \institution{Shanghai Jiao Tong University}
  \city{Shanghai}
  \country{China}
}
\email{maolinr03@sjtu.edu.cn}

\author{Xiaoyang Lu}
\affiliation{%
  \institution{Shanghai Jiao Tong University}
  \city{Shanghai}
  \country{China}
}
\email{xiaoyangl@sjtu.edu.cn}

\author{Jiaqi Liu}
\affiliation{%
  \institution{Shanghai Jiao Tong University}
  \city{Shanghai}
  \country{China}
}
\email{jkliu189@gmail.com}

\author{Jian Wang}
\affiliation{%
  \institution{CreativeFitting}
  \city{Shanghai}
  \country{China}
}
\email{jim.wang@creativefitting.ai}

\author{Weiwen Liu}
\authornotemark[1]
\affiliation{%
  \institution{Shanghai Jiao Tong University}
  \city{Shanghai}
  \country{China}
}
\email{wwliu@sjtu.edu.cn}

\author{Jianghao Lin}
\affiliation{%
  \institution{Shanghai Jiao Tong University}
  \city{Shanghai}
  \country{China}
}
\email{linjianghao@sjtu.edu.cn}

\author{Yong Yu}
\affiliation{%
  \institution{Shanghai Jiao Tong University}
  \city{Shanghai}
  \country{China}
}
\email{yyu@sjtu.edu.cn}

\author{Weinan Zhang}
\authornote{Corresponding authors.}
\affiliation{%
  \institution{Shanghai Jiao Tong University}
  \city{Shanghai}
  \country{China}
}
\email{wnzhang@sjtu.edu.cn}

\input{sections/abstract}



\maketitle

\input{sections/introduction}
\input{sections/related-work}
\input{sections/methodology}
\input{sections/experiments}
\input{sections/deployment}
\input{sections/conclusion}

\clearpage
\bibliographystyle{ACM-Reference-Format}
\bibliography{references}

\appendix
\input{sections/appendix}

\end{document}

%% file: sections/abstract.tex
\begin{abstract}
Storyboards decompose screenplays into shot-by-shot visual plans that drive automated short drama production. Because high-quality storyboarding rests on the tacit expertise of professional directors, it remains a capacity bottleneck at industrial scale. Large language models can automate this step, yet existing ways of equipping them with directorial knowledge face three challenges: (1) \emph{Knowledge acquisition}: the craft stays implicit in exemplars or must be authored by hand, so explicit knowledge exists only where a human writes it. (2) \emph{Knowledge refinement}: authored knowledge is never evaluated against execution outcomes, and opaque generation prevents feedback from being attributed to the knowledge behind each decision. (3) \emph{Knowledge injection}: injecting everything exceeds the usable context, yet hand-picking knowledge for every narrative group does not scale.
In light of these challenges, we present SAGE (\textbf{S}kill with \textbf{A}ttribution-\textbf{G}uided \textbf{E}volution). SAGE is a deployed framework that learns, attributes, and evolves directing knowledge from expert demonstrations. It first extracts content-free rules by contrasting each training screenplay with its expert storyboard. During generation, the model declares which rules each narrative group adopts. Joining these records with localized feedback yields attribution at the level of individual rules, which drives targeted rule updates. Evolved rules are then consolidated into scenario packages accessed through a routing index. Each group therefore retrieves only a bounded set of scenario packages matched to its situation, without expert intervention. On 18 test episodes across three genres, SAGE scored $77.8$ on an expert-validated rubric, exceeding professional directors' $77.1$. Deployed for 14 days in Virtual Film Studio, a commercial short drama production platform, it produced 1,344 narrative group outputs. Of these, $87.2\%$ were accepted without substantive edits, and the production team recorded a drop of over $83\%$ in authoring time per episode. We release \datasetname{}, the first public dataset pairing screenplays with storyboards authored by professional directors, spanning 68 episodes at \url{https://github.com/creDreams/PROSE}.
\end{abstract}

%% file: sections/introduction.tex
\section{Introduction}
\label{sec:intro}

\begin{figure*}[t]
  \centering
  \includegraphics[width=0.85\textwidth]{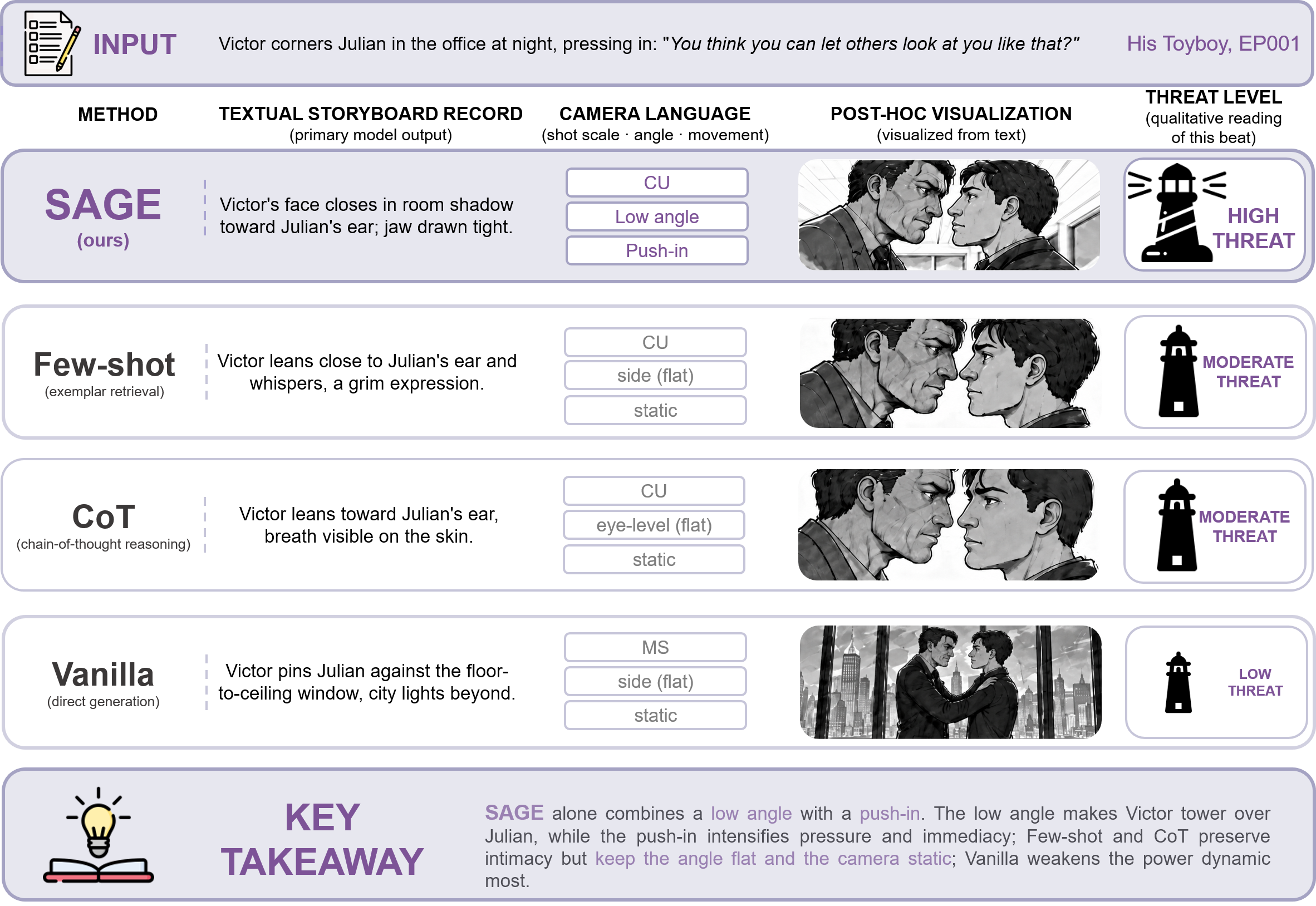}
  \caption{Camera language decisions on one screenplay beat from \emph{His Toyboy} EP001, a held-out test episode, under an identical screenplay, backbone LLM, and storyboard schema. SAGE alone selects a low-angle push-in that reinforces Victor's dominance, whereas Few-shot and CoT keep the angle flat and Vanilla widens to a composition that weakens the power dynamic. Only the textual records are system outputs; thumbnails and threat labels are post-hoc readings, excluded from evaluation.}
  \label{fig:paradigm}
  \Description{Five stacked rows compare four systems on one screenplay beat. The
  top row states the input screenplay line. The next four rows correspond to SAGE,
  Few-shot, CoT, and Vanilla. Each of these rows holds a textual storyboard record,
  three camera language fields for shot scale, camera angle, and camera movement, a
  greyscale thumbnail, and a threat level indicator. The SAGE row lists close-up,
  low angle, and push-in. The other three rows list flat angles and a static
  camera. A closing row summarizes the comparison.}
\end{figure*}

As video generation models mature, automated film production systems increasingly transform screenplays into videos~\cite{huang2025filmaster,xu2025filmagent}. In these systems, the \emph{storyboard} is the intermediate artifact connecting creative intent to video synthesis~\cite{huang2025filmaster,zhou2026dramadirector}. It specifies, shot by shot, the visual content, shot scale, camera angle, and camera movement. Its quality directly constrains the fidelity, coherence, and cinematic expressiveness of the downstream video.

Manual storyboarding constrains throughput across the short drama industry. Tens of thousands of serialized episodes ship annually on platforms such as Douyin and TikTok, yet the timeline from concept to release often spans only weeks or even days, which forces creators into multiple roles at once to meet the efficiency demands of the industry~\cite{cao2026audience}. Storyboards are still authored by hand, and this manual stage caps the speed of short drama production~\cite{wei2025cinevision,tang2024skyscript}. The bottleneck persists because storyboarding rests on tacit directorial expertise~\cite{polanyi1966tacit}. A director jointly decides shot rhythm, visual description, shot scale, camera angle, and camera movement, yet the craft behind these decisions is rarely articulated as explicit principles. On the commercial platform studied in this work, a director spends more than one hour on a single episode. Large language models (LLMs) can automate this process. Benchmarks nevertheless show that frontier models lack professional competence in camera language~\cite{liu2025shotbench,wang2025cinetechbench}. Figure~\ref{fig:paradigm} illustrates this deficit on one dramatic beat, where direct generation defaults to a flat angle that does not convey the scene's power dynamic. Effective automation therefore requires equipping LLMs with this domain knowledge.

Prior efforts inject such knowledge into LLMs through three mechanisms. Few-shot prompting supplies screenplay and storyboard exemplars, and chain-of-thought (CoT) prompting~\cite{wei2022cot} adds reasoning chains written by experts. Skills~\cite{anthropic2025skills} package a stable workflow together with separate knowledge files, a design that has become an industrial practice. These mechanisms let LLMs exploit directorial knowledge, yet three challenges remain.

\textbf{C1: Knowledge acquisition.} In existing mechanisms, knowledge is either implicit in exemplars or authored by hand. Few-shot exemplars encode the craft implicitly, so the model must re-induce it at inference time and never obtains an inspectable, reusable form. CoT chains and skill files state the knowledge explicitly, but directors must author every chain and every file themselves. Explicit knowledge is thus available only where a human writes it, and the challenge is to extract it automatically from expert demonstrations.

\textbf{C2: Knowledge refinement.} Once authored, the knowledge stays fixed and is never evaluated against execution outcomes, so its defects remain undetected. Refinement must therefore use execution feedback. Feedback alone is nevertheless insufficient, because generation is opaque: it does not reveal which piece of the injected knowledge shaped which decision. Existing pipelines judge knowledge at extraction time or by votes over whole trajectories, without tracing its use in generation~\cite{zhao2024expel,yang2023tran}, and self-correction without such localization is reported to degrade performance~\cite{huang2024selfcorrect}. Effective refinement therefore requires generation to be traceable, so that feedback can be attributed to the knowledge responsible for each decision.

\textbf{C3: Knowledge injection.} At inference time, the knowledge given to the model must be selected automatically. Current practice secures relevance by manual selection, and prior automatic retrieval still operates over knowledge units defined by hand~\cite{fu2024autoguide}. Manual selection cannot scale when every narrative group demands its own decision over thousands of entries, and injecting everything exceeds the usable context.

We present SAGE (\textbf{S}kill with \textbf{A}ttribution-\textbf{G}uided \textbf{E}volution), a framework that resolves the three challenges on the skill substrate, which already separates a stable workflow from an explicit knowledge base. For acquisition, SAGE contrasts each training screenplay with its expert storyboard and extracts content-free rules automatically (C1). For refinement, generation declares which rules each narrative group adopts. Joining these \emph{rule-adoption records} with localized feedback yields three evolution operations: misfiring rules are revised, new rules are added for coverage gaps, and unused rules are retired (C2). For injection, evolved rules are consolidated by semantic clustering into \emph{scenario packages} with a routing index, so each narrative group retrieves only the packages matched to its situation (C3).

On a test set of 18 episodes spanning three drama genres, SAGE scored $77.8$, exceeding the professional directors' $77.1$. All scores come from an LLM rubric whose agreement with three professional directors exceeds inter-human agreement (\S\ref{sec:exp:scorer}). It also outperformed strong baselines given reasoning chains authored by directors and exemplars from adjacent episodes. Attribution-guided iteration contributed $3.6$ points over rule warm-start, whereas iteration without attribution peaked early and then fell below its starting point. The consolidated rules transferred unchanged to three other backbone LLMs, with relative gains of $8.1\%$ to $21.2\%$.

Our contributions are:
\begin{itemize}
  \item \textbf{Framework.} SAGE treats the knowledge base of an LLM skill as learnable parameters and evolves it from expert demonstrations. To our knowledge, SAGE was the first framework to evolve such a knowledge base under credit assignment at the granularity of individual rules, rather than a single validation score over the whole skill document. The framework spans evolution, consolidation, and deployment, and it runs in commercial production.
  \item \textbf{Mechanism.} A rule-level attribution mechanism that assigns credit over knowledge expressed in natural language. It links authoring to execution feedback, a connection that static skills lack. Our ablation and iteration studies show that iteration without it peaks early and then declines.
  \item \textbf{Dataset.} We release \datasetname{}, to our knowledge the first open dataset pairing screenplays with storyboards \emph{authored by professional human directors}\footnote{Publicly available at \url{https://github.com/creDreams/PROSE}.}. It spans 68 episodes across three professionally produced series of distinct genres. Prior resources instead provide shot annotations that models reverse-engineer from finished videos~\cite{tang2024skyscript,zhou2026dramadirector}.
  \item \textbf{Deployment.} We deployed SAGE in Virtual Film Studio (VFS), a commercial short drama production platform, and report a 14-day study on three unseen ongoing dramas. Of 1,344 narrative group outputs, $87.2\%$ entered production without substantive edits. Authoring time per episode fell from over one hour to roughly 10 minutes, a reduction of over $83\%$. Storyboarding therefore shifted from manual authoring to review, which is what removes the capacity bottleneck.
\end{itemize}

%% file: sections/related-work.tex
\section{Related Work}
\label{sec:related}

\textbf{LLM-based storyboard generation.}
FilmAgent~\cite{xu2025filmagent} and MovieAgent~\cite{wu2025movieagent} coordinate role-playing agents. FilMaster~\cite{huang2025filmaster} instead retrieves camera language conventions from 440K film clips. Neither learns an explicit, inspectable knowledge base from professional storyboard demonstrations. DramaDirector~\cite{zhou2026dramadirector} is the closest system to our task, since it fine-tunes an LLM planner with SFT and GRPO for short drama storyboards, but it keeps cinematic knowledge implicit in model weights. A parallel line generates \emph{image} storyboards~\cite{xie2024storyboard20k,dinkevich2025story2board}, addressing visual consistency rather than the directorial decomposition studied here, and earlier engine-based previsualization renders shot candidates under \emph{manually specified} rules~\cite{rao2023vds}. Benchmark studies consistently find that frontier models lack professional competence in camera language~\cite{liu2025shotbench,wang2025cinetechbench}, which motivates explicit knowledge injection.

\textbf{Storyboard datasets.}
SkyScript-100M~\cite{tang2024skyscript} and DramaBoard~\cite{zhou2026dramadirector} pair short drama scripts with shot-level annotations. Their storyboards, however, are \emph{reverse-engineered from finished videos}, so they capture what ended up on screen rather than the decisions the director made. \datasetname{} instead releases the directors' original pre-production storyboards, the decision traces that demonstration-supervised evolution requires.

\textbf{Experiential knowledge in self-evolving agents.}
Self-evolving agents~\cite{gao2025survey} improve from their own experience; we focus on the branch that evolves context rather than weights. Self-Refine~\cite{madaan2023selfrefine} and Reflexion~\cite{shinn2023reflexion} iterate on individual outputs or store trajectory-level reflections, without accumulating knowledge that persists beyond the task. Unaided self-correction is also known to be unreliable~\cite{huang2024selfcorrect}. A second family distills experience into persistent natural-language knowledge, expressed as failure-derived rules, state-conditioned guidelines, causal abstractions, or distilled insights and procedural memory~\cite{yang2023tran,fu2024autoguide,majumder2023clin,zhao2024expel,fang2025memp}. A third family stores capabilities as executable skills. Voyager~\cite{wang2023voyager} grows a code skill library through environment feedback, and successors extend the idea to other interactive domains~\cite{tan2024cradle,zheng2025skillweaver,wang2024awm}; industrial standards package such procedural instructions with resources~\cite{anthropic2025skills}, and memory architectures manage the state generically~\cite{packer2023memgpt,zhang2024memorysurvey}. Across this family the maintenance signal is coarse: items are added or voted on from trajectory outcomes, without tracking \emph{which} item influenced \emph{which} output. Sound and harmful items therefore receive the same credit, and the noise grows as the store accumulates entries. AutoManual~\cite{chen2024automanual} comes closest, since its planner cites the rules it engages, but it learns from binary episodic rewards and revises rules by post-hoc judgment over whole trajectories. SAGE instead aligns against expert demonstrations, the only supervision available without an environment that verifies success, and joins per-group adoption records with localized feedback (\S\ref{sec:exp:iteration}). Agent-Pro~\cite{zhang2024agentpro} and AgentEvolver~\cite{zhai2025agentevolver} also use reflection or self-attribution, but operate over policy prompts and RL action steps rather than declarative knowledge items.

\textbf{Natural-language component optimization.}
Manually supplied context is brittle, since in-context learning is sensitive to exemplar relevance and ordering, and selecting effective exemplars is itself a retrieval problem~\cite{liu2022makesgood,zhao2021calibrate,lu2022fantastically,rubin2022learning}. Treating the natural-language components of a frozen-LLM system as optimizable parameters began with instruction and evolutionary prompt search~\cite{zhou2023ape,yang2024opro,fernando2024promptbreeder,guo2024evoprompt}. The idea then matured into gradient-like frameworks. ProTeGi~\cite{pryzant2023protegi} edits prompts along critique-derived textual gradients, while TextGrad~\cite{yuksekgonul2024textgrad} and Trace~\cite{cheng2024trace} backpropagate language feedback through computation graphs and execution traces. DSPy~\cite{khattab2024dspy} compiles declarative pipelines with learnable instructions, and later frameworks train agent functions as weights~\cite{zhang2024agentoptimizer} or show that language-level reflection can outperform RL in sample efficiency~\cite{agrawal2025gepa}. Most recently the paradigm has reached skills. EvoSkill~\cite{alzubi2026evoskill} edits skill folders from failure analysis, and SkillOpt~\cite{yang2026skillopt} treats skill documents as trainable state. In both, the learning signal remains a single scalar validation score over the whole skill, so edits are retained or discarded in bulk and an individual rule's contribution is never measured. SAGE shares the view of knowledge as learnable parameters, but maintains rule-adoption records that assign credit to individual rules, so one round can revise, add, and retire different rules at once. Our experiments show that this granularity keeps iteration productive where document-level optimization plateaus (\S\ref{sec:exp:main}).

%% file: sections/methodology.tex
\section{Methodology}
\label{sec:method}

We present SAGE, a framework that enables the knowledge component of an LLM skill to evolve autonomously from expert demonstrations.

\subsection{Problem Formulation}
\label{sec:method:formulation}

\paragraph{Storyboard generation.}
A screenplay $S$ consists of scenes with dialogue, action descriptions, and character information. The goal is to produce a storyboard $B = \langle b_1, \ldots, b_n \rangle$, an ordered sequence of shots. Each shot $b_i$ is a structured tuple specifying visual content, shot scale, camera angle, camera movement, characters, and dialogue. Operationally, each scene is partitioned into \emph{narrative groups}; every group yields a \emph{storyboard segment} of one or more shots, and these segments are merged into $B$. Producing $B$ requires joint decisions across five professional dimensions: \emph{shot rhythm}, \emph{visual description}, \emph{shot scale}, \emph{camera angle}, and \emph{camera movement}. These dimensions encode tacit directorial knowledge that is difficult to specify exhaustively in a static prompt.

\paragraph{Skill as workflow and knowledge.}
We define a \emph{skill} as a pair $\Sigma = (\mathcal{W}, \mathcal{R})$. The \emph{workflow} $\mathcal{W}$ is a fixed multi-step procedural scaffold. It specifies how the screenplay is partitioned, which knowledge is retrieved at each step, and how partial outputs are merged. The \emph{knowledge base} $\mathcal{R}$ holds the declarative rules that the workflow consumes. This decomposition follows emerging industrial standards for agent skills~\cite{anthropic2025skills}. Each rule $r \in \mathcal{R}$ is a pair
\begin{equation}
r = (\textit{cond}, \textit{prac}),
\end{equation}
where \textit{cond} describes the narrative situation in which the rule applies, such as a shock reaction within a high-intensity dialogue. The second element \textit{prac} is an executable directive, such as not inserting a breathing shot between that shock reaction and the follow-up question. Each rule is tagged with one of the five dimensions. Rules are further constrained to be \emph{content-free}: they must not contain concrete shot content or verbatim material from expert storyboards, which prevents data leakage and encourages generalization.

\paragraph{Learning view.}
The workflow $\mathcal{W}$ is easy to fix by design, whereas skill quality is dominated by $\mathcal{R}$. We therefore cast rule acquisition as optimization. Let $g_{\Sigma}(S)$ denote the storyboard generated by an LLM equipped with skill $\Sigma$. Let $f_{\mathrm{align}}(g_{\Sigma}(S), B^{*}) \in [0, 100]$ measure how closely that storyboard matches the expert reference $B^{*}$ across the five dimensions. Given a corpus of expert demonstrations $\mathcal{D} = \{(S_j, B_j^{*})\}_{j=1}^{m}$, training seeks
\begin{equation}
\mathcal{R}^{\star} = \arg\max_{\mathcal{R}} \; \mathbb{E}_{(S, B^{*}) \sim \mathcal{D}} \left[ f_{\mathrm{align}}\big(g_{(\mathcal{W}, \mathcal{R})}(S), B^{*}\big) \right].
\end{equation}
This formulation yields a direct analogy to standard machine learning. The rule set $\mathcal{R}$ plays the role of \emph{learnable parameters}, and $f_{\mathrm{align}}$ acts as the \emph{training objective}. The \emph{test metric} is a separate reference-free quality score $f_{\mathrm{qual}}$, which measures absolute professional quality without access to $B^{*}$. Two properties distinguish this setting from gradient-based learning. The ``parameters'' are discrete natural-language rules, and the optimization signal must be routed to individual rules through an explicit \emph{attribution} mechanism (\S\ref{sec:method:evolution}).

\subsection{Framework Overview}
\label{sec:method:overview}

\begin{figure*}[t]
  \centering
  \includegraphics[width=\textwidth]{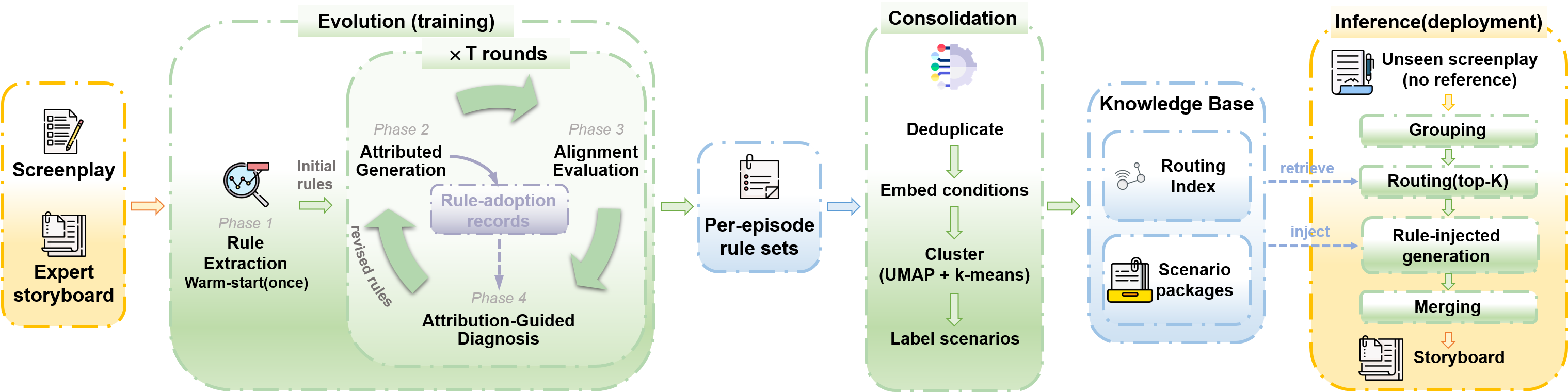}
  \caption{Overview of SAGE. \textbf{Stage 1 (Evolution)}: for each training episode, a four-phase loop generates storyboards with per-group rule attribution, evaluates them against the expert reference, and revises the rule set through attribution-guided diagnosis. \textbf{Stage 2 (Consolidation)}: rules evolved across all episodes are deduplicated, embedded, and clustered into scenario packages with a routing index. \textbf{Stage 3 (Inference)}: on unseen screenplays, narrative groups are routed to their matched scenario packages, whose rules are injected into generation; no expert reference is required.}
  \label{fig:framework}
  \Description{A horizontal pipeline of four blocks. The leftmost block holds the
  screenplay and the expert storyboard. The Evolution block encloses a cycle of
  four numbered phases, namely rule extraction, attributed generation, alignment
  evaluation, and attribution-guided diagnosis, with rule-adoption records at the
  center and a loop labeled T rounds. The Consolidation block lists deduplication,
  condition embedding, clustering, and scenario labeling. The Knowledge Base block
  holds a routing index and scenario packages. The rightmost Inference block lists
  grouping, top-k routing, rule-injected generation, and merging into a
  storyboard.}
\end{figure*}

As shown in Figure~\ref{fig:framework}, SAGE operates in three stages. All three share a unified generation pipeline of \emph{narrative grouping}, \emph{scenario routing}, \emph{rule injection}, \emph{attributed generation}, and \emph{segment merging}. \textbf{Stage~1 (evolution, \S\ref{sec:method:evolution})} refines a per-episode rule set through a four-phase loop whose key ingredient is \emph{rule-level attribution}. \textbf{Stage~2 (consolidation, \S\ref{sec:method:consolidation})} deduplicates and clusters the large, redundant union of per-episode rule sets into \emph{scenario packages} with a routing index. \textbf{Stage~3 (inference, \S\ref{sec:method:inference})} routes each group of an unseen screenplay to its top-$k$ packages and generates a segment with the retrieved rules injected.

\subsection{Stage 1: Attribution-Guided Rule Evolution}
\label{sec:method:evolution}

\begin{figure}[t]
  \centering
  \includegraphics[width=\columnwidth]{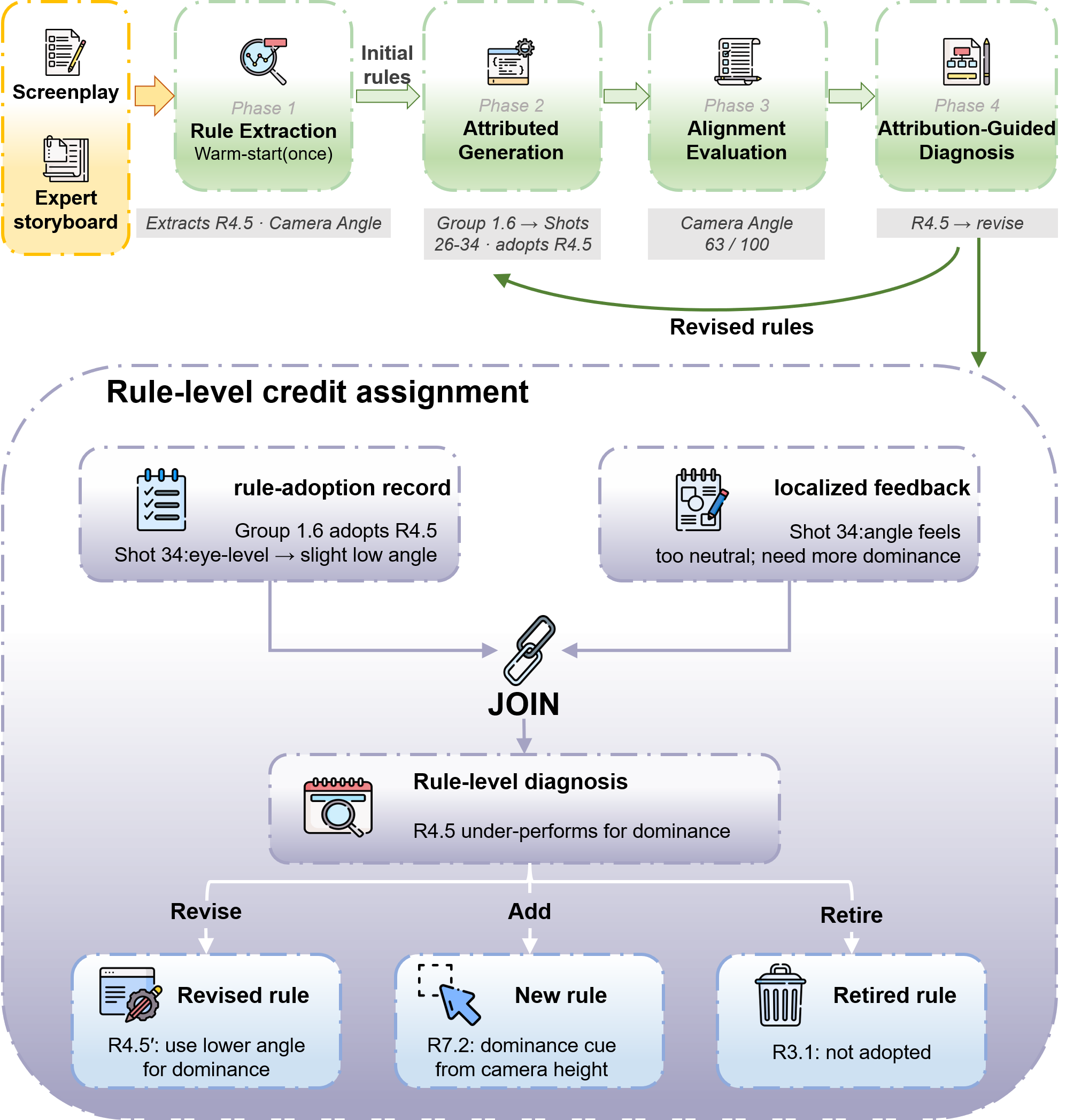}
  \caption{One round of attribution-guided rule evolution, on a real trace from \emph{Beyond the Wall}. Attributed generation (Phase 2) records which rules each group adopted; diagnosis (Phase 4) joins these records with dimension-level feedback to decide, per rule, whether to revise, add, or retire.}
  \label{fig:evolution-loop}
  \Description{The upper part shows the four phases of one evolution round in
  sequence, annotated with a real trace in which a camera angle rule scores 63 out
  of 100 and is marked for revision. The lower part expands rule-level credit
  assignment. A rule-adoption record and a localized feedback item meet at a join
  operation, which produces a rule-level diagnosis. Three outcomes branch from the
  diagnosis, namely a revised rule, a newly added rule, and a retired rule.}
\end{figure}

The evolution stage refines the rule set for each training episode over $T$ rounds. Each round executes the four phases shown in Figure~\ref{fig:evolution-loop}, and Algorithm~\ref{alg:evolution} summarizes the loop.

\paragraph{Phase 1: Rule Extraction.}
In the first round, an initial rule set $\mathcal{R}^{(1)}$ is extracted by contrasting screenplay with expert storyboard. The screenplay is first partitioned into a two-level hierarchy, in which scenes at level~1 are split into \emph{narrative groups} at level~2. A group is a dialogue exchange, an action sequence, or an emotional beat, and is the minimal unit of analysis. For every group, the extractor examines how the director decomposed it into shots, then induces content-free rules $(\textit{cond}, \textit{prac})$ that explain the observed decisions. Subsequent rounds inherit the revised rule set from the previous round's Phase~4.

\paragraph{Phase 2: Attributed Generation.}
The model generates one segment per group from three inputs. These are the screenplay, a director vocabulary defining the legal shot scales, angles, and movements, and the current rule set $\mathcal{R}^{(t)}$. The expert storyboard is withheld. The defining feature of this phase is the \emph{rule-adoption record}: for each group $u$, the model declares the adopted rules $A_u \subseteq \mathcal{R}^{(t)}$ alongside the segment it produces. The full \emph{attribution map} $\mathcal{A}^{(t)} = \{(u, A_u)\}_{u \in \mathrm{groups}(S)}$ makes every generation decision traceable to the rules that informed it.

\paragraph{Phase 3: Alignment Evaluation.}
The storyboard is scored against the expert reference by $f_{\mathrm{align}}$, which produces an overall score, five per-dimension scores, and natural-language feedback that localizes each deviation. One such deviation reads that the confrontation in scene~4 lacks a re-establishing two-shot after three consecutive close-ups. The expert storyboard is visible only to the evaluator, never to the generator. Appendix~\ref{app:align} details the review protocol.

\paragraph{Phase 4: Attribution-Guided Diagnosis.}
Diagnosis joins the feedback with $\mathcal{A}^{(t)}$ to perform rule-level credit assignment. Deviations fall into two classes with distinct remedies. In \textbf{Class~A (misfiring rule)}, a dimension deviates in groups where some rule $r$ \emph{is} adopted. That rule is implicated, so its condition or practice is \emph{revised}. In \textbf{Class~B (coverage gap)}, a dimension deviates in groups where \emph{no} adopted rule governs it. No existing rule is at fault, so a \emph{new} rule is induced from the feedback. Rules never adopted throughout the episode are \emph{retired}, which keeps the set minimal. At most $30\%$ of rules may be modified per round, a cap that prevents destructive oscillation. The revised set $\mathcal{R}^{(t+1)}$ seeds the next round.

Without attribution, feedback can only be assigned at the episode level. The optimizer then knows \emph{that} a dimension scored poorly but not \emph{which} rule caused it, so revisions become undirected rewrites. Our ablations identify this shift from episode-level to rule-level credit assignment as the condition for sustained improvement (\S\ref{sec:experiments}).

\begin{algorithm}[b]
\caption{Attribution-Guided Rule Evolution for a Single Episode}
\label{alg:evolution}
\linespread{1.24}\selectfont
\begin{algorithmic}[1]
\REQUIRE screenplay $S$, expert storyboard $B^{*}$, rounds $T$
\ENSURE evolved rule set $\mathcal{R}^{(T+1)}$
\STATE $U \leftarrow \mathrm{Group}(S)$ \COMMENT{two-level narrative grouping}
\STATE $\mathcal{R}^{(1)} \leftarrow \mathrm{ExtractRules}(S, B^{*}, U)$ \COMMENT{Phase 1}
\FOR{$t = 1$ \TO $T$}
  \STATE $(B^{(t)}, \mathcal{A}^{(t)}) \leftarrow \mathrm{AttrGen}(S, U, \mathcal{R}^{(t)})$ \COMMENT{Phase 2}
  \STATE $(s^{(t)}, F^{(t)}) \leftarrow f_{\mathrm{align}}(B^{(t)}, B^{*})$ \COMMENT{Phase 3}
  \FORALL{deviations $d \in F^{(t)}$}
    \IF{$\exists\, r \in A_u$ governing $\dim(d)$ for the group $u$ of $d$}
      \STATE revise $r$ \COMMENT{Class A: misfiring rule}
    \ELSE
      \STATE $\mathcal{R}^{(t)} \leftarrow \mathcal{R}^{(t)} \cup \{\mathrm{InduceRule}(d)\}$ \COMMENT{Class B: gap}
    \ENDIF
  \ENDFOR
  \STATE retire rules never adopted in $\mathcal{A}^{(t)}$
  \STATE $\mathcal{R}^{(t+1)} \leftarrow$ revised set \COMMENT{$\leq 30\%$ of rules modified}
\ENDFOR
\end{algorithmic}
\end{algorithm}

\subsection{Stage 2: Rule Consolidation}
\label{sec:method:consolidation}

Evolution is per-episode by design. The union across the corpus reaches the order of $10^2$ rules per episode and several thousand in total, and is redundant and in places contradictory. Consolidation compresses it into a retrievable knowledge base in three steps.

\paragraph{Deduplication.}
A two-level union-find procedure runs within each dimension. At level~1, rules whose \emph{condition} embeddings exceed a cosine similarity of $0.90$ are merged into a condition group. At level~2, within each condition group, rules whose \emph{practice} embeddings also exceed the threshold are collapsed to their semantic centroid. Practices below the threshold are preserved as alternative practices of a single multi-practice rule. One pass thus resolves both redundancy, where condition and practice coincide, and latent contradiction, where a shared condition maps to divergent practices. The threshold is deliberately conservative because the two error directions are not symmetric. Merging rules that differ in meaning destroys knowledge irrecoverably, whereas failing to merge equivalent rules only leaves redundancy, since divergent practices survive as alternatives of one rule instead of being collapsed into a centroid.

\paragraph{Embedding and clustering.}
Each deduplicated rule is represented by its condition embedding, capturing the narrative situation it targets. Embeddings are $\ell_2$-normalized, reduced with UMAP~\cite{mcinnes2018umap}, and clustered with $k$-means. A grid search selects the number of clusters and the UMAP hyperparameters, jointly scoring dimension coverage, cluster size compliance, size uniformity, and silhouette quality. The number of scenario packages is therefore determined by the data rather than fixed a priori. Rules triggered by similar situations thus become co-located and co-retrieved, regardless of their source episode or drama.

\paragraph{Scenario packaging.}
An LLM agent labels each cluster with a human-readable scenario name, such as \emph{emotional climax under psychological pressure}, together with a short applicability description. The agent then materializes the cluster as a \emph{scenario package}, a document that groups the cluster's rules by dimension. A compact \emph{routing index} is built alongside, listing every package's name, description, and rule inventory. Appendix~\ref{app:consolidation} shows the resulting cluster structure and an example package.

\subsection{Stage 3: Scenario-Aware Inference}
\label{sec:method:inference}

At deployment the evolved skill runs on unseen screenplays with no expert reference and no iteration, reusing the training pipeline. The screenplay is partitioned into the same two-level hierarchy used during evolution. For each group, the model matches its situation against the routing index and selects the top-$k$ packages, recording a justification per match. We set $k{=}3$ to match the three reference episodes supplied to Few-shot and CoT, which equalizes the injection budget across knowledge-injection methods. Routing over the index rather than scanning all rules keeps the injected context bounded as the knowledge base grows. Each group's segment is then generated independently and in parallel, conditioned on the group's screenplay content, the retrieved packages, and the director vocabulary. Generation also emits a rule-adoption record, which preserves traceability in deployment. Finally, segments are concatenated in screenplay order and their shots renumbered into the final storyboard. Because packages encode situation-conditioned knowledge rather than model-specific tricks, the consolidated base is backbone-agnostic and can be injected into other LLMs unchanged. Appendix~\ref{app:routing} traces one real group through routing and generation.

%% file: sections/experiments.tex
\section{Experiments}
\label{sec:experiments}

We evaluate SAGE around four research questions. \textbf{(RQ1)} Does the evolved skill close the quality gap to professional directors, and how does it compare with strong prompting and skill optimization baselines? \textbf{(RQ2)} How much does each component contribute, namely rule warm-start, iteration, and attribution? \textbf{(RQ3)} Does attribution make quality improve over rounds instead of fluctuating? \textbf{(RQ4)} Is the consolidated knowledge base portable across backbone LLMs?

\subsection{Experimental Setup}
\label{sec:exp:setup}

\paragraph{Dataset.}
\datasetname{} comprises three professionally produced short drama series of distinct genres, namely a sci-fi suspense series of 20 episodes (\emph{Beyond the Wall}), an urban romance series of 23 (\emph{His Toyboy}), and an emotional healing series of 25 (\emph{My Cure}). Each episode pairs a screenplay, comprising a synopsis, character profiles, and a scene-level script, with the storyboard authored by the series' professional director. We held out 6 episodes per series, 18 in total, as the test set. The consolidated knowledge base is built exclusively from rules evolved on the remaining 50 training episodes.

\paragraph{Evaluation protocol.}
All systems were scored by a reference-free \emph{quality} rubric on the five dimensions of shot rhythm, visual description, shot scale, camera angle, and camera movement. Each dimension uses a 100-point scale, and the overall score is their average. Scoring used Claude Opus~4.6 under a fixed rubric prompt, whose score anchors are given in Appendix~\ref{app:quality}. This metric is distinct from the alignment score $f_{\mathrm{align}}$ used as the training signal: quality measures how \emph{good} a storyboard is against professional standards, whereas alignment measures how \emph{close} it is to a specific expert reference. The scorer's reliability is validated against human experts in \S\ref{sec:exp:scorer}.

\paragraph{Baselines.}
We compared six alternatives under the same backbone (Claude Opus~4.6) and output schema. \textbf{Director} is the human storyboard, which serves as the expert reference. \textbf{Vanilla} generates directly with no external knowledge. \textbf{Few-shot} is conditioned on $\langle$screenplay, storyboard$\rangle$ pairs from the \emph{three nearest neighboring episodes of the same series}, never the target episode itself. \textbf{CoT} adds reasoning chains that encode decomposition thinking \emph{authored by directors} on those same reference episodes. \textbf{EvoSkill}~\cite{alzubi2026evoskill} and \textbf{SkillOpt}~\cite{yang2026skillopt} are representative skill optimization methods, reimplemented faithfully on the same test set. The strong baselines thus receive demonstrations from adjacent episodes, whereas SAGE uses none at inference, which makes the comparison conservative for our method.

\paragraph{Implementation.}
Rule evolution ran $T{=}10$ rounds per training episode with at most $30\%$ of rules modified per round. Conditions were embedded with Qwen3-Embedding-8B; consolidation yielded $55$ scenario packages from $2{,}036$ deduplicated rules. Inference routed each group to its top-3 packages. Unless stated otherwise, SAGE results use the round-5 knowledge base, which is the best-performing round on the test set. The no-attribution ablation is likewise reported at its own best round (\S\ref{sec:exp:iteration}), so this oracle round selection is applied symmetrically and characterizes each variant's upper bound.

\subsection{Main Results (RQ1)}
\label{sec:exp:main}

\begin{table}[t]
\caption{Main comparison on the 18-episode test set (quality scores, 100-point scale). \textbf{Bold}: best among AI systems; \underline{underline}: exceeds the human director.}
\label{tab:main}
\centering
\small
\begin{tabular}{@{}lcccccc@{}}
\toprule
Method & Rhythm & Visual & Scale & Angle & Move. & Overall \\
\midrule
Director & 80.1 & 75.1 & \textbf{81.7} & \textbf{76.8} & \textbf{71.6} & 77.1 \\
\midrule
Vanilla & 68.2 & 67.4 & 74.4 & 62.1 & 53.8 & 65.2 \\
Few-shot & 73.1 & 73.3 & 78.2 & 68.0 & 64.1 & 71.2 \\
CoT & \textbf{79.4} & 80.8 & 79.5 & 72.1 & 68.4 & 76.0 \\
EvoSkill & 73.9 & 72.2 & 74.3 & 62.8 & 59.7 & 68.7 \\
SkillOpt & 78.9 & 74.9 & 79.5 & 74.4 & 68.1 & 75.1 \\
SAGE (ours) & 79.2 & \underline{\textbf{85.1}} & 78.8 & 74.6 & 71.2 & \underline{\textbf{77.8}} \\
\bottomrule
\end{tabular}
\end{table}

Table~\ref{tab:main} reports the main comparison, from which three findings emerge. \emph{Expert-level quality.} At $77.8$ overall, SAGE was the only AI system to exceed the human director at $77.1$. It was also the closest system to the director on camera angle and camera movement, the two dimensions on which Vanilla scored lowest. \emph{Knowledge versus exemplars.} Both prompting baselines received demonstrations from adjacent episodes and reasoning authored by directors, yet Few-shot reached only $71.2$ and CoT $76.0$. SAGE encodes the same knowledge \emph{explicitly}, instead of leaving it latent in exemplars for the model to induce anew. \emph{Generic skill optimization.} EvoSkill and SkillOpt both scored below SAGE, with their largest deficits on camera movement, the dimension with the lowest scores overall. SAGE thus gained most where Vanilla was weakest, and its visual description even surpassed the director. We attribute this to evolved rules that enforce compositional completeness in lighting, blocking, and framing, which human storyboards often leave implicit.

\subsection{Ablation Study (RQ2)}
\label{sec:exp:ablation}

\begin{table}[t]
\caption{Ablation on the 18-episode test set. Each row adds one component.}
\label{tab:ablation}
\centering
\small
\begin{tabular}{@{}llcc@{}}
\toprule
& Configuration & Overall & $\Delta$ \\
\midrule
A & Vanilla (no external knowledge) & 65.2 & $-$ \\
B & + rule warm-start (no iteration) & 74.2 & $+9.0$ \\
C & + iteration (no attribution) & 75.4 & $+1.2$ \\
D & + attribution (full SAGE) & \textbf{77.8} & $+2.4$ \\
\bottomrule
\end{tabular}
\end{table}

Table~\ref{tab:ablation} isolates each component. The contrastive warm-start in row~B provided the largest single gain, since rules extracted by contrasting screenplays with expert storyboards already capture substantial explicit knowledge. Iteration \emph{without} attribution in row~C added little, because episode-level feedback cannot identify which rules to fix. Adding attribution in row~D more than doubled the iteration benefit, with its largest gains on shot scale and visual description, the two dimensions where row~C remained weakest. This pattern confirms the argument of \S\ref{sec:method:evolution}: rule-level credit assignment converts iteration from perturbation into optimization.

\subsection{Iteration Dynamics (RQ3)}
\label{sec:exp:iteration}

\begin{figure}[t]
  \centering
  \includegraphics[width=\columnwidth]{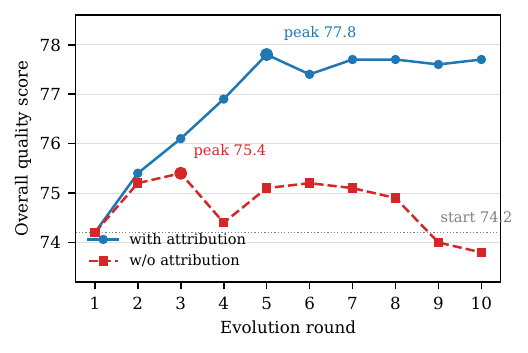}
  \caption{Quality on the test set over 10 evolution rounds. Both settings share the same round-1 rule set at 74.2. With attribution, quality rises to 77.8 by round 5 and stays within 77.4 to 77.7 through round 10; without attribution, it peaks at 75.4 in round 3 and degrades to 73.8 by round 10.}
  \label{fig:iteration}
  \Description{A line chart with evolution round from 1 to 10 on the horizontal
  axis and overall quality score from 73 to 78 on the vertical axis. Both lines
  start together at 74.2, marked by a dotted horizontal reference line. The solid
  line for the setting with attribution climbs to a peak of 77.8 at round 5 and
  then stays close to that level. The dashed line for the setting without
  attribution peaks at 75.4 at round 3 and then declines to 73.8, ending below the
  starting level.}
\end{figure}

Figure~\ref{fig:iteration} tracks quality across 10 rounds from an identical warm-start set. With attribution, quality rose over the first five rounds and then held stable through round 10. Without attribution, it peaked earlier at a lower value and ended \emph{below its starting point}, so attribution changes both the magnitude and the stability of improvement.

\subsection{Cross-Model Generalization (RQ4)}
\label{sec:exp:crossmodel}

\begin{figure}[b]
  \centering
  \includegraphics[width=\columnwidth]{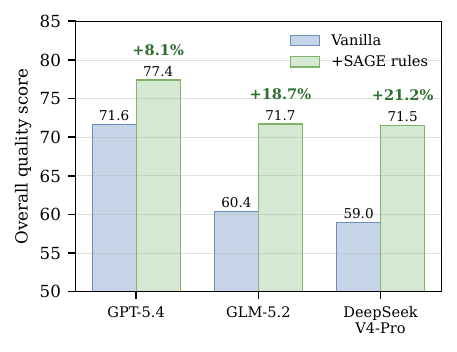}
  \caption{Cross-model transfer of the consolidated knowledge base, evolved entirely with Claude Opus 4.6. Injecting the unchanged scenario packages improves every transfer target; labels above the SAGE bars report relative improvements over Vanilla.}
  \label{fig:crossmodel}
  \Description{A grouped bar chart over three backbone models, namely GPT-5.4,
  GLM-5.2, and DeepSeek V4-Pro. Each model has a Vanilla bar and a bar for the same
  model with SAGE rules injected. The Vanilla scores are 71.6, 60.4, and 59.0. The
  scores with SAGE rules are 77.4, 71.7, and 71.5. Relative improvements of 8.1
  percent, 18.7 percent, and 21.2 percent are printed above the second bar of each
  group.}
\end{figure}

If the evolved rules encode directorial domain knowledge rather than backbone-specific tricks, they should transfer to other LLMs unchanged. We injected the identical scenario packages, evolved entirely with Claude Opus~4.6, into three backbones and modified nothing else in the pipeline. As Figure~\ref{fig:crossmodel} shows, every target improved by $8.1\%$ to $21.2\%$ relative, and GPT-5.4 approached the source model's own quality.\footnote{GLM-5.2 was evaluated on 16 of the 18 episodes owing to two routing failures, with the matching Vanilla episodes excluded for parity.} Two patterns are notable. First, weaker backbones benefited more, because the rules supply structure that compensates for missing domain knowledge, whereas the strongest backbone gained least from the highest baseline. Second, visual description transferred most universally, which makes compositional checklists the most portable evolved knowledge. Absolute scores nonetheless tracked backbone capability, since rules supply domain knowledge but not generation capability.
\subsection{Validity of the Automatic Scorer}
\label{sec:exp:scorer}

\begin{table}[b]
\caption{Scorer validity on 18 director storyboards: agreement of Claude Opus 4.6 with the consensus of three professional directors, vs.\ inter-human agreement, measured by Lin's CCC.}
\label{tab:scorer}
\centering
\small
\begin{tabular}{@{}lcc@{}}
\toprule
Dimension & Claude vs.\ human consensus & Inter-human \\
\midrule
Shot rhythm & \textbf{0.717} & 0.689 \\
Visual description & \textbf{0.893} & 0.664 \\
Shot scale & 0.674 & \textbf{0.708} \\
Camera angle & \textbf{0.873} & 0.809 \\
Camera movement & \textbf{0.873} & 0.627 \\
\midrule
Mean & \textbf{0.806} & 0.699 \\
\bottomrule
\end{tabular}
\end{table}

All reported scores come from an LLM scorer, a paradigm whose reliability and biases are well documented~\cite{zheng2023judging,panickssery2024llm}. We therefore validated it against human judgment. Three professional directors and the scorer independently scored the 18 director storyboards on the five dimensions, yielding 90 score pairs. Table~\ref{tab:scorer} reports Lin's concordance correlation coefficient~\cite{lin1989ccc}. The scorer's agreement with the human consensus \emph{exceeded} inter-human agreement on four of the five dimensions, the sole exception being shot scale. Human scores were on average lower by a small margin, a systematic offset that does not affect relative rankings. Self-preference bias~\cite{panickssery2024llm} is also unlikely to favor our method, since all AI systems in Table~\ref{tab:main} share the scorer's backbone and any such bias applies uniformly. We conclude the scorer is a reliable proxy for expert judgment in this domain.

%% file: sections/deployment.tex
\section{Production Deployment}
\label{sec:deployment}

CreativeFitting is an AI native entertainment company based in Shanghai. It operates Reel.AI, among the first AI generated short drama apps distributed to overseas audiences on the App Store and Google Play, and VFS, its in-house creation platform on which over a thousand creators produce content. To test whether offline gains translate into production value, we deployed SAGE in VFS, using the same framework trained on a larger proprietary corpus of director demonstrations. The evaluation covered three ongoing productions disjoint from the public 68-episode corpus. Over 14 days, platform logs recorded 12 production users, 1,344 narrative group outputs, and 2,038 generation and revision operations.

We computed acceptance at the narrative group level, the unit of independent generation. Following the production team's operational criterion, an output is accepted if it can enter downstream production without substantive edits, where changes limited to asset references, formatting, punctuation, or wording count as non-substantive.

\begin{table}[h]
\caption{Production acceptance on three ongoing dramas. Each output corresponds to one narrative group and is a segment that may contain multiple shots.}
\label{tab:deployment}
\centering
\small
\begin{tabular}{@{}lrrrr@{}}
\toprule
& Prod. A & Prod. B & Prod. C & Overall \\
\midrule
Narrative group outputs & 559 & 453 & 332 & 1,344 \\
Accepted outputs & 460 & 380 & 332 & 1,172 \\
Acceptance (\%) & 82.3 & 83.9 & 100.0 & \textbf{87.2} \\
\bottomrule
\end{tabular}
\end{table}

Table~\ref{tab:deployment} shows that 1,172 of 1,344 outputs were accepted without substantive edits, an overall rate of $87.2\%$ that ranged from $82.3\%$ to $100.0\%$ across the three productions. The production team further reported that typical authoring time per episode fell from over one hour to roughly 10 minutes, an approximately sixfold acceleration. The acceptance rate is computed from platform interaction logs, whereas the turnaround was tracked by the production team over the same period.

Two properties of the deployed pipeline keep the residual manual effort bounded. The unit of acceptance coincides with the unit of generation, so a rejected output calls for a local regeneration of one narrative group rather than a revision pass over the episode. The system also emits rule-adoption records at inference time (\S\ref{sec:method:inference}), so every rejected output stays traceable to the rules that informed it.

%% file: sections/conclusion.tex
\section{Conclusion}
\label{sec:conclusion}

Professional storyboarding depends on directorial knowledge that experts cannot exhaustively articulate, so every existing injection path relies on manual externalization. SAGE removes this dependence by evolving the knowledge component of a skill from the expert demonstrations released in \datasetname{}. Rule-adoption records route feedback to individual rules, and the evolved rules are consolidated into scenario packages for deployment without a reference. The evolved knowledge exceeded the professional directors on our test set, transferred unchanged to three other backbones, and held these gains in a production deployment on live dramas.

Our iteration study also generalizes beyond storyboarding. The \emph{granularity} of credit assignment determines whether knowledge evolution converges: rule-level attribution reached a stable optimum, whereas episode-level feedback declined. Systems that treat natural-language knowledge as learnable parameters therefore need to localize feedback to individual items. This requirement is architectural rather than domain specific, since any pipeline whose generation step declares the knowledge it consumed can route feedback to that knowledge.

%% file: sections/appendix.tex
\section{Rule Consolidation Details}
\label{app:consolidation}
\begin{figure}[H]
  \centering
  \begin{subfigure}[b]{0.42\columnwidth}
    \includegraphics[width=\textwidth]{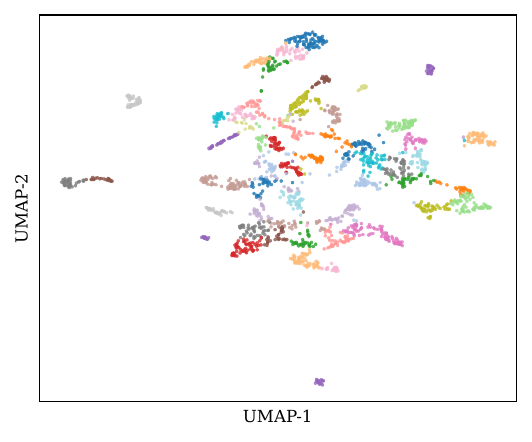}
    \caption{}
    \label{fig:consolidation:a}
  \end{subfigure}\hfill
  \begin{subfigure}[b]{0.45\columnwidth}
    \includegraphics[width=\textwidth]{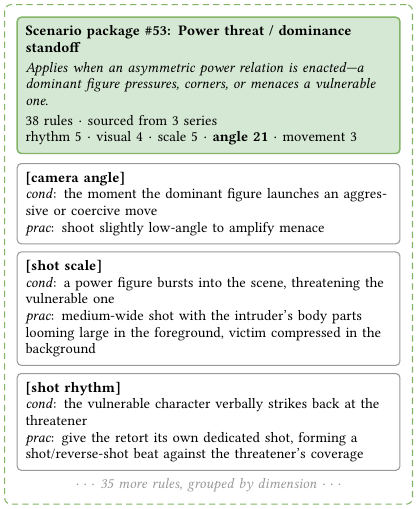}
    \caption{}
    \label{fig:consolidation:b}
  \end{subfigure}
  \caption{Rule consolidation. (a) UMAP projection of rule-condition embeddings, colored by cluster. (b) A representative scenario package shown in translation: a routable natural-language description over rules grouped by professional dimension. The package contains 38 rules in total; representative rules are shown for space.}
  \label{fig:consolidation}
  \Description{Panel (a) is a two-dimensional UMAP scatter plot of rule condition
  embeddings. Points form many small compact groups, each drawn in a distinct
  color, separated by empty space. Panel (b) shows one scenario package as a
  document. A header names the package and gives its applicability description, the
  rule count of 38, the three contributing series, and the rule counts per
  dimension. Three framed entries follow, one each for camera angle, shot scale,
  and shot rhythm, and each entry states a condition and a practice. A footer notes
  that 35 further rules are omitted.}
\end{figure}

Figure~\ref{fig:consolidation} shows that rules learned across episodes and series form coherent, well-separated clusters (a). Each cluster becomes a self-contained scenario package, routable by its natural-language description and organized by professional dimension (b). This is how evolved knowledge is stored and retrieved at inference time (\S\ref{sec:method:consolidation}).

Two properties are worth noting. First, clusters mix rules from all three series rather than separating by source, indicating that the learned conditions describe narrative situations rather than one drama's idiosyncrasies. The example package contains 38 rules contributed by all three series, and the recovered scenario carries no trace of its source episodes. Second, the distribution across dimensions is intentionally uneven: camera angle dominates this power-asymmetry package, whereas an emotional-release package concentrates on shot scale and rhythm. Consolidation therefore preserves each situation's dimensional signature rather than balancing dimensions artificially.

\section{Scenario-Aware Inference Example}
\label{app:routing}
\begin{figure}[H]
  \centering
  \includegraphics[width=0.72\columnwidth]{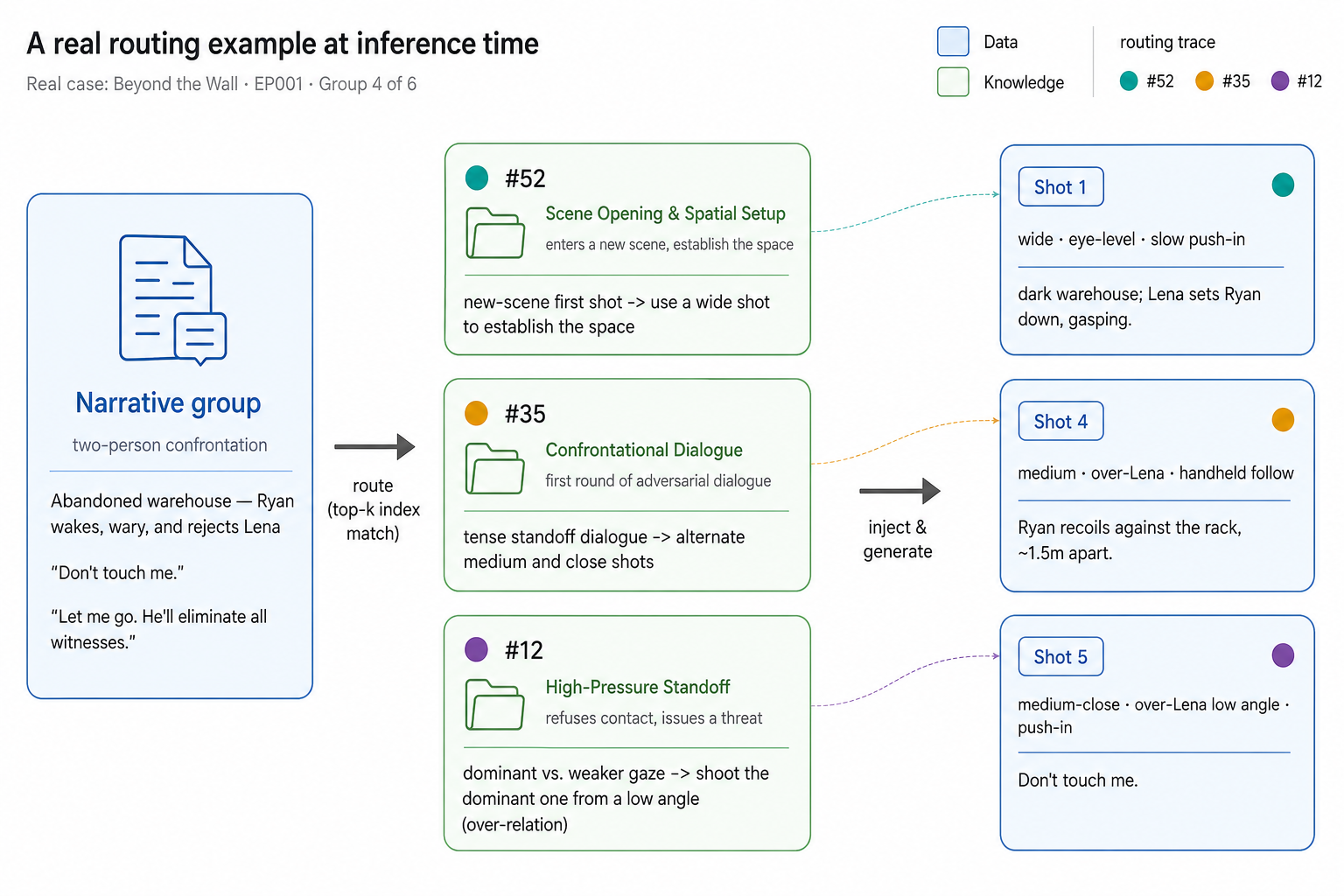}
  \caption{A real routing example from \emph{Beyond the Wall} EP001, a held-out test episode, translated. Each shot is marked with the \emph{primary} package whose rules it adopted, and one representative rule per package is shown.}
  \label{fig:routing-example}
  \Description{A left to right flow in three columns. The left column describes one
  narrative group, a two-person confrontation, with its dialogue lines. The middle
  column lists the three scenario packages it was routed to, each with a
  identifier, a name, a condition, and a practice. The right column shows the
  generated shots, each labeled with its shot scale, camera angle, and camera
  movement. Colored connecting lines trace which package informed which shot.}
\end{figure}

Figure~\ref{fig:routing-example} traces one real narrative group from a test episode through the scenario-aware inference pipeline of \S\ref{sec:method:inference}. A two-person confrontation is matched against the routing index and dispatched to its top-3 scenario packages, whose rules are injected into generation; the produced shots then carry rule-adoption records back to the packages that informed them. No expert reference is involved.

The three retrieved packages are complementary rather than redundant. One supplies the angle vocabulary for power asymmetry, another governs the rhythm of a verbal exchange, and the third covers the framing of a physical intrusion. Their rules therefore act on different shots of the same segment, and the adoption records make this division visible after the fact, which is what allows a rejected output to be traced to the rule that shaped it.

\clearpage

\section{Alignment Review Skill}
\label{app:align}

The alignment score $f_{\mathrm{align}}$ of \S\ref{sec:method:formulation} is produced by a review skill that compares a generated storyboard against the director reference. Figure~\ref{fig:skill-align} reproduces that skill file, translated and condensed to fit the column. The reviewer receives both artifacts, whereas the generator never sees the reference. Comparison proceeds over the five professional dimensions, and the director storyboard is treated as the sole correct target on every one of them.

Two design choices make the resulting signal usable for rule level diagnosis. First, the skill discards all surface variation. Table layout, column order, and wording are excluded from the comparison, so a deviation reflects a directorial decision rather than a formatting artifact. Second, the report records only deviations. Praise and hedging are suppressed, and each deviation must cite the shot numbers at which the two storyboards diverge. A deviation without such a citation cannot be attributed and is therefore rejected. Phase~4 of evolution consumes these localized deviations and joins them with the rule adoption records, as described in \S\ref{sec:method:evolution}.

\begin{figure}[H]
  \centering
  \input{figures/tikz_skill_alignment}
  \caption{The skill file behind $f_{\mathrm{align}}$, translated and condensed to fit the column. Section headings and the wording of every directive follow the original. The director storyboard is the gold standard and stays hidden from the generator.}
  \label{fig:skill-align}
  \Description{A framed transcript of the alignment review skill file, set in a
  monospaced font under a colored title bar. Sections appear in order. A role
  section casts the reader as a storyboard review expert. A review dimensions
  section enumerates the five professional dimensions and instructs the reviewer to
  ignore table format, column order, and wording. Further sections state the
  scoring procedure and the report format.}
\end{figure}

\newpage

\section{Quality Scoring Rubric}
\label{app:quality}

The quality metric $f_{\mathrm{qual}}$ of \S\ref{sec:exp:setup} scores a storyboard on absolute professional standards without any reference. Figure~\ref{fig:skill-qual} reproduces the scoring prompt, translated and condensed to fit the column. Its grounding is the editing priority of Walter Murch, which ranks emotion and story above rhythm and continuity~\cite{murch2001blink}. Every dimension is read in the context of vertical short drama, where narrative density is high and the opening seconds govern retention.

Each dimension carries a 100 point scale divided into five bands of twenty points, and the overall score is their unweighted mean. Every band is anchored by the observable evidence expected at that level rather than by an adjective, which keeps runs comparable. The bands share one structure across the five dimensions, so a score reads the same wherever it appears. Three protocol rules govern the output. Every credit or deduction cites specific shot numbers. Judgment follows short drama practice instead of feature film convention. Section~\ref{sec:exp:scorer} validates the rubric against three professional directors.

\begin{figure}[H]
  \centering
  \input{figures/tikz_skill_quality}
  \caption{The scoring prompt behind $f_{\mathrm{qual}}$, translated and condensed to fit the column. The five bands are shown for the rhythm dimension; the other four share the same structure and are abridged to their core criteria.}
  \label{fig:skill-qual}
  \Description{A framed transcript of the quality scoring prompt, set in a
  monospaced font under a colored title bar. A role section casts the reader as a
  senior storyboard director and states that no reference storyboard is available.
  A grounding paragraph lists Murch's six criteria for editing with their weights,
  the function of a storyboard, and the traits of short drama. A score anchors
  section then gives five bands of twenty points for the rhythm dimension, each band
  stating the observable evidence expected at that level.}
\end{figure}

%% file: figures/tikz_skill_alignment.tex
\skillbox{skAlign}{Alignment Review Skill \normalfont(condensed from the skill file used for $f_{\mathrm{align}}$)}{%
\textcolor{skAlign}{\bfseries\# Role}\\[1.2pt]
You are a professional storyboard review expert. Compare each AI storyboard
against the director's original storyboard and score its degree of alignment.\\[3pt]
\textcolor{skAlign}{\bfseries\# Review dimensions}\\[1.2pt]
Follow the five dimensions below. \textcolor{skWarn}{Ignore table format, column order,
and wording entirely.}\\[1.5pt]
1. \textbf{Rhythm} --- shot splitting density, timing of inserted reaction shots,
emotional breathing room.\\[0.8pt]
2. \textbf{Visual description} --- purely visual content, physical action, micro
expression and physiological reaction such as a swallow or a tremble.
Ignore all interiority and literary embellishment.\\[0.8pt]
3. \textbf{Shot scale} --- scale selection, including the director's preference for
close-up and extreme close-up on body detail.\\[0.8pt]
4. \textbf{Camera angle} --- subjective and objective viewpoint shifts, over the
shoulder framing, Dutch angle, and the power dynamics built by low and
high angles.\\[0.8pt]
5. \textbf{Camera movement} --- visual stability of the move, such as a locked off
frame or a slow push in.\\[3pt]
\textcolor{skAlign}{\bfseries\# Scoring}\\[1.2pt]
Each dimension uses a 100-point scale; the overall score is their mean.
A higher score means closer alignment with the director.
\textcolor{skWarn}{The director's manual storyboard is the only correct
alignment target (gold standard).}\\[3pt]
\textcolor{skAlign}{\bfseries\# Workflow}\\[1.2pt]
1. Ask the user for the paths of the AI storyboards and of the director
benchmark file before starting.\\[0.8pt]
2. Write a Python script to read the CSV files and analyse the differences.\\[0.8pt]
3. Never print a whole table, which overflows the context. Slice the data or
search for action keywords to locate the dramatic peaks, then compare
reaction close-ups and oppressive compositions there.\\[0.8pt]
4. Emit the report in the two prescribed parts.\\[3pt]
\textcolor{skAlign}{\bfseries\# Schema pitfalls}\\[1.2pt]
The two sources carry different headers, so resolve each field through a
cascade of fallbacks. The AI output merges scale, viewpoint, and composition
into one column, whereas the director file splits scale and angle apart. Read
every file as \texttt{utf-8-sig} so that a byte order mark cannot corrupt the
first header. Guard against generated files that hold a header but no rows.\\[3pt]
\textcolor{skAlign}{\bfseries\# Output}\\[1.2pt]
\textbf{Part 1.} A summary table: one row per plan, with the episode, the overall
score, the five dimension scores, and a one-line justification.\\[1.5pt]
\textbf{Part 2.} A deviation analysis, under strict discipline:\\[0.8pt]
$\bullet$ \textcolor{skWarn}{Pain points only. Never state an advantage of the AI plan,
a weakness of the director, or any word of praise.}\\[0.8pt]
$\bullet$ \textbf{Cite shot numbers.} For example: in Shot 15 the director uses a
Dutch low angle with an over-the-shoulder framing to convey Victor's
pressure, whereas plan A stays at eye level and loses the spatial
hierarchy entirely.\\[0.8pt]
$\bullet$ Report explicitly whether the plan misses the director's preferred
visual grammar, the handheld breathing quality, or the listener's
reaction shot.%
}

%% file: figures/tikz_skill_quality.tex
\skillbox{skQual}{Quality Scoring Rubric \normalfont(condensed from the prompt used for $f_{\mathrm{qual}}$)}{%
\textcolor{skQual}{\bfseries\# Role}\\[1.2pt]
You are a senior storyboard director and cinematographer with twenty years of
experience, fluent in vertical short drama production. Score the given
storyboard on five dimensions, 100 points each, \textcolor{skWarn}{with no
reference storyboard available.}\\[1.5pt]
Ground every judgment in: \textbf{Murch's six criteria for editing} --- emotion
51\%, story 23\%, rhythm 10\%, eye trace 7\%, planarity 5\%, spatial continuity
4\%; the \textbf{function of a storyboard} as the route map from script to
screen, where every frame carries a narrative purpose; and \textbf{short drama
traits} --- dense narration, mobile first, the first seconds deciding
retention, a twist every 30 to 60 seconds.\\[3pt]
\textcolor{skQual}{\bfseries\# Score anchors} \normalfont(five bands of twenty points, the
same structure on every dimension; the rhythm dimension is shown)\\[1.5pt]
\textbf{81--100} complete rhythmic arc of setup, escalation, climax, and
breath; density gradient precisely matched to tension; ASL near 1.5 to 2s at
the climax and 3 to 5s in dialogue; deliberate breathing shots; the first
three shots form an effective hook.\\[0.8pt]
\textbf{61--80} rhythm varies plausibly and the gradient is broadly right;
breathing shots exist but sit imprecisely.\\[0.8pt]
\textbf{41--60} basic fast and slow variation, yet mechanical; breathing shots
absent or misplaced; the opening lacks a hook.\\[0.8pt]
\textbf{21--40} scattered variation with no density logic; action and dialogue
are barely distinguished.\\[0.8pt]
\textbf{0--20} monotonous throughout; no breathing shot; the climax is
indistinguishable from the setup, and shots are merely enumerated.\\[3pt]
\textcolor{skQual}{\bfseries\# Core criteria per dimension} \normalfont(abridged)\\[1.5pt]
\textbf{Rhythm} --- density gradient, breathing design, hook structure, arc
within a scene, motivation for each cut.\\[0.8pt]
\textbf{Visual description} --- executable composition, light direction and
quality, spatial layering, precision of character description, environmental
storytelling, purely visual content only.\\[0.8pt]
\textbf{Shot scale} --- spectrum coverage from ELS to ECU, emotional weight
matching, progression logic, the 30-degree rule, restraint on ECU.\\[0.8pt]
\textbf{Camera angle} --- narrative motivation, power dynamics through low and
high angles, the 180-degree axis, POV placement, restraint on Dutch angle.\\[0.8pt]
\textbf{Camera movement} --- trigger, path, and stop point for every move;
static against moving contrast; semantics of direction; type variety.\\[1.5pt]
\textcolor{skWarn}{Vertical rules override film convention:} 9:16 framing, a
naturally high share of CU and MCU, and push-in to CU as the signature reveal.\\[3pt]
\textcolor{skQual}{\bfseries\# Output format}\\[1.2pt]
Report a score table over the five dimensions with a grade each and the
overall mean. Grades follow the bands: excellent, good, fair, pass, fail.\\[1.5pt]
Then, per dimension, give the reasoning with shot numbers as evidence, the
strongest case, the main defect, and a concrete suggestion targeting it. Close
with a verdict naming the single most important strength and weakness.\\[3pt]
\textcolor{skQual}{\bfseries\# Cautions}\\[1.2pt]
Judge by the standard of a working short drama director rather than relaxing
it because the plan already beats most AI output.
\textcolor{skWarn}{Every deduction must cite shot numbers.} Stay in the short
drama context and never import feature film standards. Executability comes
first: the bottom line is whether a crew could shoot the plan without further
discussion. A few standout shots do not offset a systemic defect, since
consistency across the episode matters more than a local highlight. \textcolor{skWarn}{The five dimensions must
remain discriminative, so do not assign one score to all of them.}%
}